\documentclass{article}

\usepackage[preprint]{neurips_2026}

\usepackage[utf8]{inputenc} 
\usepackage[T1]{fontenc}    
\usepackage{hyperref}       
\usepackage{url}            
\usepackage{booktabs}       
\usepackage{amsfonts}       
\usepackage{nicefrac}       
\usepackage{microtype}      
\usepackage{xcolor}         
\usepackage{tabularx,multirow}
\usepackage{makecell}
\usepackage{graphicx}
\usepackage{amsmath}
\usepackage{graphicx}
\usepackage{subcaption}
\usepackage{booktabs}
\usepackage{wrapfig}
\usepackage[normalem]{ulem}
\useunder{\uline}{\ul}{}
\usepackage{amssymb}
\usepackage{float}
\usepackage{placeins}
\DeclareUnicodeCharacter{03B2}{\ensuremath{\beta}}
\DeclareUnicodeCharacter{03BE}{\ensuremath{\xi}}
\setcitestyle{numbers,square}

\title{$\xi$-DPO: Direct Preference Optimization via Ratio Reward Margin}

\author{%
  Zhengyuan~Fan$^{1}$ \quad
Zhonghua~Wu$^{1}$ \quad
Yuxuan~Du$^{1}$ \quad
Qun~Chen$^{1}$\thanks{Corresponding author.} \\
$^{1}$School of Computer Science, Northwestern Polytechnical University \\
\texttt{\{fanzhengyuan,wuxhua,duyuxuan36,chenbenben\}@nwpu.edu.cn}
}

\begin{document}
\raggedbottom

\maketitle

\begingroup
\renewcommand\thefootnote{}
\footnotetext{Code is available at \url{https://github.com/zyfan1/Xi-DPO}.}
\endgroup

	\begin{abstract}
		Reference-free preference optimization has emerged as an efficient alternative to reinforcement learning from human feedback, with Simple Preference Optimization(SimPO) demonstrating strong performance by eliminating the explicit reference model through a simple objective. However, the joint tuning of the hyperparameters $\beta$ and $\gamma$ in SimPO remains a central challenge. We argue that this difficulty arises because the margin formulation in SimPO is not easily interpretable across datasets with different reward gap structures. To better understand this issue, we conduct a comprehensive analysis of SimPO and find that $\beta$ implicitly controls sample filtering, while the effect of $\gamma$ depends on the reward gap structure of the dataset. Motivated by these observations, we propose $\xi$-DPO: Direct preference optimization via ratio reward margin. We first reformulate the preference objective through an equivalent transformation, changing the optimization target from maximizing the likelihood of reward gaps to minimizing the distance between reward gaps and optimal margins. Then, we redefine the reward in a ratio form between the chosen and rejected, which effectively cancels the effect of $\beta$ and yields a bounded and interpretable margin. This margin is called the ratio reward margin and is denoted by $\xi$. {Explicitly representing the desired relative separation between chosen and rejected responses, $\xi$ can be determined from the initial reward gap distribution, thus much easier to tune than the $\beta$ and $\gamma$ in SimPO.} Finally, our extensive experiments show that $\xi$-DPO substantially outperforms existing preference optimization methods across open benchmarks on multiple evaluation metrics.
	\end{abstract}

	\section{Introduction}
	
	With the rapid advancement of large language models\cite{qwen3, gpt5}, aligning their responses with human preferences has become critically important. Ouyang et al.\cite{RLHF} introduced reinforcement learning from human feedback (RLHF), a method for aligning large language model outputs with human preferences. In their framework, the model is optimized using the Proximal Policy Optimization (PPO) algorithm\cite{PPO}. RLHF comprises three stages: 1. Supervised fine-tuning(SFT) of large models on downstream tasks; 2. Reward modeling based on the SFT model; 3. The final reinforcement learning stage. Despite its effectiveness, this multi-stage pipeline introduces considerable complexity into the training process. 
	
	Christiano et al.\cite{rlhf2} propose using the Bradley-Terry model(BT model)\cite{bt} for preference modeling to optimize the reward model, $r(y, x)$: Given a dataset $D=\{(x, y_w, y_l)\}$, it consists of prompts $x$ and paired responses ($y_w$, $y_l$), where $y_w$ is the chosen response(winning) and $y_l$ is the rejected response(losing). Their preference relationship is
	\begin{equation}
		\begin{aligned}
			p(y_w \succ y_l \mid x)
			&= \frac{\exp\big(r(y_w, x)\big)}{\exp\big(r(y_w, x)\big)+\exp\big(r(y_l,x)\big)}
			\\
			&= \sigma\big(r(y_w, x)-r(y_l, x)\big)
			\label{eq:bt_first}
		\end{aligned}
	\end{equation}
	where $\sigma$ is sigmoid function. Subsequent work, such as Direct Preference Optimization (DPO)\cite{DPO}, significantly simplifies RLHF. The key innovation of DPO lies in deriving the relationship between reward model and optimization policy, thereby merging reward modeling and reinforcement learning into one single stage. This significantly simplifies the process of preference optimization. Researchers only need to optimize the target model $\pi_\theta$ using the following DPO loss function to align the model’s responses with human preferences:
	\begin{equation}
		\begin{aligned}
			&
			\ell_{\mathrm{DPO}}(\theta)
			= \mathbb{E}_{(x,y_w,y_l) \sim D}\!\Biggl[\\
			&
			-\log \sigma\!\left(
			\beta \left(
			\log \frac{\pi_\theta(y_w|x)}{\pi_{\mathrm{ref}}(y_w|x)}
			-
			\log \frac{\pi_\theta(y_l|x)}{\pi_{\mathrm{ref}}(y_l|x)}
			\right)
			\right)
			\Biggr].
		\end{aligned}
	\end{equation}
	where $\pi_\theta$ is the policy model to be optimized, $\pi_{\mathrm{ref}}$ is the reference policy, $\pi_\theta(y|x)$ and $\pi_{\mathrm{ref}}(y|x)$ denote the probabilities of generating response $y$ given prompt $x$ under the policy model and the reference model respectively, and $y_w$ denotes the chosen response and $y_l$ the rejected response.

	Recent studies\cite{beta, matters} have shown that the DPO hyperparameter $\beta$ is highly sensitive to the data distribution, which makes tuning difficult and can result in only marginal performance gains. Wu et al.\cite{beta} analyzed how the choice of $\beta$ depends on the data distribution through experiments. Specifically, when the reward gap between the chosen ($y_w$) and rejected ($y_l$) responses is large, a larger $\beta$ is preferred; when the gap is small, $\beta$ should be smaller.  Simple Preference Optimization(SimPO)\cite{simpo} provides a more efficient formulation by rewriting the reward form of DPO\cite{DPO} $\beta\log \frac{\pi_\theta(y\mid x)}{\pi_{\mathrm{ref}}(y\mid x)}$ to $\frac{\beta}{|y|}log \pi_\theta(y|x)$ and further encourages that the reward gap between the chosen response $y_w$ and the rejected response $y_l$ is at least $\gamma$. They highlighted the necessity of $\gamma$, arguing that it directly affects the uniformity, or flatness of the reward-gap distribution. Formally, the loss function of SimPO is defined as:
	\begin{equation}
		\begin{aligned}
			&
			\ell_{\mathrm{SimPO}}(\theta)
			=
			\mathbb{E}_{(x,y_w,y_l) \sim \mathcal{D}}
			\Biggl[\\
			&
			-\log \sigma
			\left(
			\frac{\beta}{|y_w|} \log \pi_\theta(y_w \mid x)
			-
			\frac{\beta}{|y_l|} \log \pi_\theta(y_l \mid x)
			-
			\gamma
			\right)
			\Biggr].
			\label{eq:simpo_}
		\end{aligned}
	\end{equation}
	
	SimPO requires the joint tuning of $\beta$ and $\gamma$. As in $\beta$-DPO\cite{beta}, we use a small, fixed $\beta$ for the low-gap data and a larger $\beta$ for the high-gap data. We fine-tune Pythia-2.8B\cite{pythia} with SimPO\cite{simpo}, setting $\gamma$ to decay from large to small(6 to 1). The detailed GPT-4 based evaluation results are presented in Table~\ref{tab:winrate_gamma}. It can be observed that the win rate on the low-gap data increases from 3.2\% to 49.5\%, while on the high-gap data, the win rate decreases from 31.46\% to 30.16\%. As shown in Table \ref{tab:winrate_gamma}, both $\beta$ and $\gamma$ exhibit high sensitivity, making it necessary to conduct multiple experiments to identify appropriate values. These results also reveal an interpretability issue regarding the reward margin $\gamma$ in SimPO. In our experiments on $\gamma$, varying $\gamma$ leads to substantially larger performance changes on low-gap data than on high-gap data.  From this viewpoint, $\gamma$ is regarded as a constraint strength that matches the intrinsic gap structure of the data, rather than as a margin expected to force the reward gaps of most samples beyond a prescribed threshold. Based on our analysis, when SimPO-like methods are applied across different datasets, the intrinsic reward gaps of these datasets vary. However, such variation is difficult to quantify, making it unclear whether $\gamma$ should be increased or decreased, and by how much. This uncertainty is one of the main reasons why selecting an appropriate $\gamma$ is challenging.

	\begin{table*}[t]
		\centering
		\caption{\textbf{Win rate variations with $\beta$, $\gamma$ across different data types.} The left table presents the results from $\beta$-DPO\cite{beta}, illustrating how the effect of $\beta$ varies across different data types. The right table reports our results, exploring how the effect of $\gamma$ changes with the data type.}
		\label{tab:winrate_gamma}
		\begin{minipage}{0.465\textwidth}
			\centering
			\caption*{(a) Win rate variations with $\beta$. The results from $\beta$-DPO.}
			\begin{tabular}{lccc}
				\toprule
				Data type & $\beta=0.1$ & $\beta=0.3$ & $\beta=0.5$ \\
				\midrule
				Low gap  & 43.0 & 37.0 & 33.0 \\
				High gap & 7.0  & 28.0 & 31.0 \\
				\bottomrule
			\end{tabular}
		\end{minipage}
		\hfill
		\begin{minipage}{0.50\textwidth}
			\centering
			\caption*{(b) Win rate variations with $\gamma$. The results are obtained from our experiments built upon SimPO.}
			\begin{tabular}{lcccc}
				\toprule
				Data type & $\beta$ & $\gamma=6$ & $\gamma=3$ & $\gamma=1$ \\
				\midrule
				Low gap  & 2  & 3.20  & 46.94 & 49.45 \\
				High gap & 10 & 31.46 & 30.70 & 30.16 \\
				\bottomrule
			\end{tabular}
		\end{minipage}
	\end{table*}

	Both DPO and SimPO adopt an optimization objective of the form $\sigma(\beta(r(y_w|x)- r(y_l|x)))$. The $\beta$ typically acts as a scaling factor. However, in this paper, we revisit the role of $\beta$ from a different perspective and show that it implicitly serves to filter out positive high-gap samples. It provides a compelling explanation for why existing dynamic hyperparameter tuning strategies\cite{beta, aDPO} are effective in practice. Based on these insights, we propose $\xi$-DPO: Direct preference optimization via ratio reward margin. First, we simplify the optimization objective of SimPO through an equivalence function mapping. It transforms the goal from maximizing the probability likelihood of reward gaps into minimizing the mean squared error between reward gap and the theoretically optimal gap. Second, we normalize the reward by converting it into a ratio of the chosen to the rejected. This normalization not only effectively cancels out $\beta$, but also constrains the reward gap to the interval [-1, 1], which we define as the ratio reward margin $\xi$. Finally, we employ LeakyReLU activation function to prevent reward degradation, where reward gaps that already exceed $\xi$ are pulled back, leading to an increase in the rejected reward and a decrease in the chosen reward. 
	
	Formally, the optimization objective of $\xi$-DPO is defined as follows: 
	
	\begin{equation}
		\begin{aligned}
			&
			\min_{\theta}\;
			\mathbb{E}_{(x,y_w,y_l)\sim\mathcal{D}}
			\Biggl[\\
			&
			\operatorname{LeakyReLU}
			\left(
			\xi
			-
			\left(\frac{
				\frac{1}{|y_w|}\log \pi_{\theta}(y_w \mid x)
				-
				\frac{1}{|y_l|}\log \pi_{\theta}(y_l \mid x)
			}{
				\left|
				\frac{1}{|y_w|}\log \pi_{\theta}(y_w \mid x)
				+
				\frac{1}{|y_l|}\log \pi_{\theta}(y_l \mid x)
				\right|
			}\right)
			\right)^2
			\Biggr]
			\label{eq:our_loss_first}
		\end{aligned}
	\end{equation}
	where $\operatorname{LeakyReLU}$ is activate function. {The above equation has only one adjustable parameter, $\xi$, which explicitly represents the desired relative separation between chosen and
		rejected responses. Unlike the $\beta$ and $\gamma$ in SimPO, $\xi$ can be selected based on the initial reward gap distribution rather than careful trial-and-error tuning.} In practical implementations, $\xi$ is supposed to be determined by the quantiles of the initial reward gap distribution of the policy. Specifically, for low-gap datasets constructed using a strong reward model, we suggest setting $\xi$ within the 90th--95th percentile range of the distribution. This choice allows most samples to participate in training while preventing excessively strong reward signals from causing the model to overfit. For high-gap datasets constructed using a weaker reward model, typically within the 97th--99.9th percentile, which helps avoid premature termination of optimization caused by insufficient reward signals. Our sensitivity experiments further show that $\xi$-DPO remains robust to the choice of $\xi$ as long as it is selected within a reasonable range. 


	
	
	\begin{figure*}[t]
		\centering
		\begin{subfigure}[b]{0.48\textwidth}
			\centering
			\includegraphics[width=\textwidth]{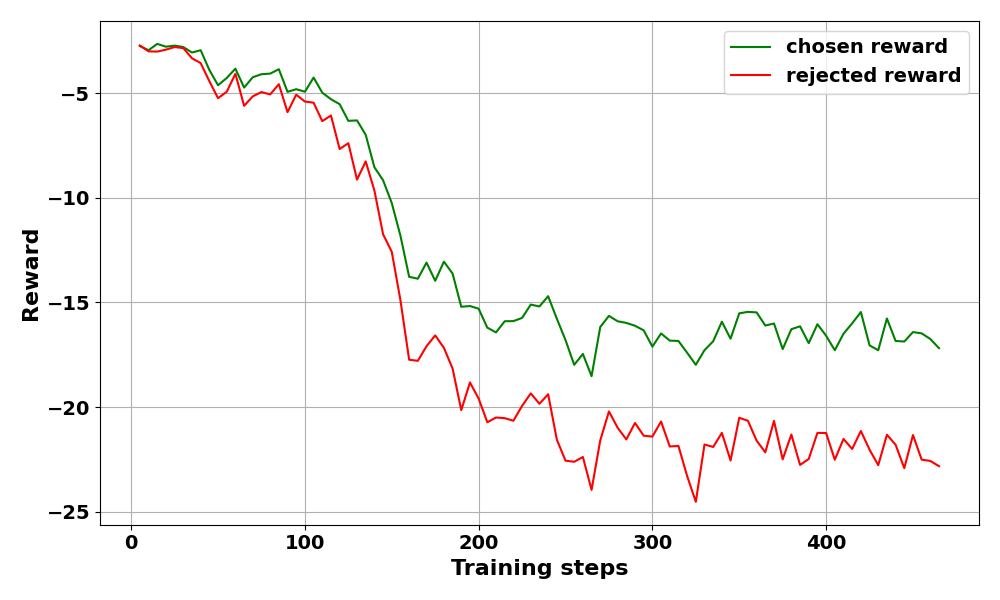}
			\caption{}
			\label{fig:chosen}
		\end{subfigure}
		\hfill
		\begin{subfigure}[b]{0.48\textwidth}
			\centering
			\includegraphics[width=\textwidth]{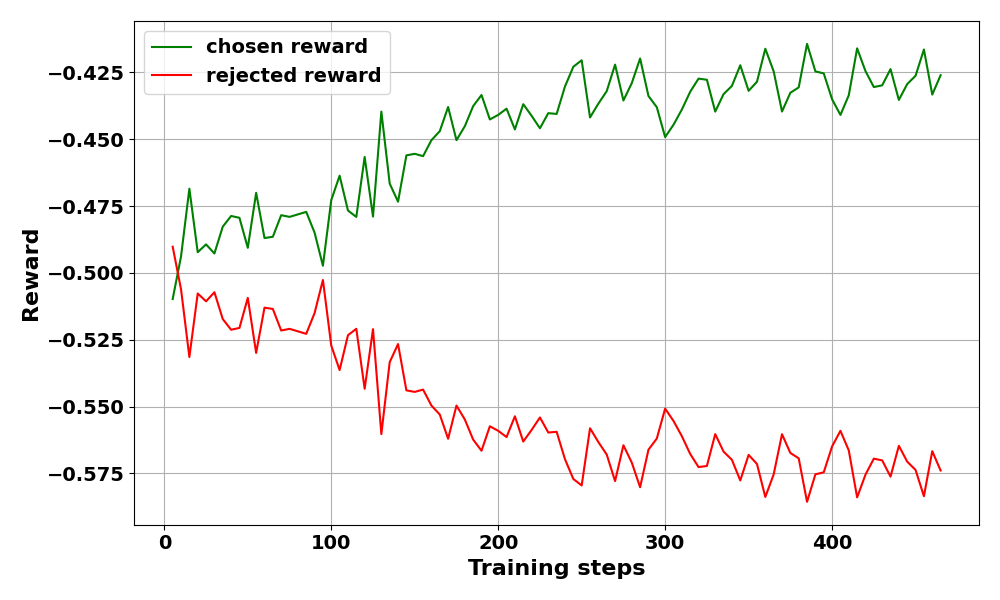}
			\caption{}
			\label{fig:rejected}
		\end{subfigure}
		\caption{Comparison of reward curves. (a) shows the reward dynamics of AlphaDPO during training, and (b) presents those of $\xi$-DPO. For $\xi$-DPO, the reward for the chosen response steadily increases while that for the rejected response decreases, indicating that the model becomes increasingly aligned with the chosen response, in accordance with our optimization objective. By contrast, for AlphaDPO method such as SimPO, although the chosen reward remains higher than the rejected reward, both display a downward trend, implying a weaker ability to separate chosen responses from rejected ones.}
		\label{fig:reward_compare}
	\end{figure*}

	It is noteworthy that the three designs of $\xi$-DPO are key to its efficacy. \textbf{The equivalence mapping} of optimization objective eliminates the adverse optimization effects caused by the hyperparameter of $\beta$ on the log-sigmoid gradient, while endowing the reward margin with an explicit semantic role of enforcing the distinction between chosen and rejected responses. \textbf{The reward redefinition} cancels out $\beta$ and makes the reward margin bounded. Finally, \textbf{LeakyReLU} safeguards this enforced separation: The reward of high-gap samples is not forcibly pulled back. We visualize the reward curves during the training processes of $\xi$-DPO and AlphaDPO, a dynamic-$\gamma$ variant of SimPO\cite{aDPO}, in Figure \ref{fig:reward_compare}. As shown in the figure, for the AlphaDPO methods such as SimPO, although the reward of the chosen remains higher than that of the rejected, both display a downward trend, implying a weaker ability to separate chosen responses from rejected ones. In contrast, for $\xi$-DPO, the reward of the chosen steadily increases while the reward of the rejected decreases, indicating that the model becomes increasingly aligned with the chosen response, in accordance with our optimization objective.

	We summarize our contributions as follows:
	
	\begin{enumerate}
		\item we systematically analyze the roles of the hyperparameters, $\beta$ and $\gamma$, in SimPO and arrive at two insights: i) $\beta$ is not only a reward-scaling factor, but also serves to filter out positive high-gap samples; ii) from a token-level perspective, $\gamma$ is determined by the intrinsic reward gap of the data. The uncertainty introduced by $\beta$ in sample filtering and the difficulty of quantifying the intrinsic reward gaps of different datasets, lead to the sensitivity of $\beta$ and $\gamma$, making them difficult to select.
		
		\item Based on the insights from our $\beta$ and $\gamma$ role analysis, we propose $\xi$-DPO. It adopts a simpler optimization objective with only one hyperparameter of $\xi$, which can be easily set based on dataset features rather than careful trial-and-error tuning. Its novel structure, together with the margin derived from the quantiles of the reward-gap distribution, also effectively mitigates hyperparameter sensitivity. 
		
		\item Extensive experiments across multiple SFT models and preference optimization benchmarks demonstrate the effectiveness and generality of $\xi$-DPO, showing that it achieves superior performance compared with existing alternative methods. Our sensitivity evaluation also reveals that the performance of $\xi$-DPO is robust w.r.t $\xi$, provided that its value is set 
		within a reasonable range. 
	\end{enumerate}

	\section{Related Work}
	\textbf{Reinforcement Learning from Human Feedback.} RLHF demonstrates effectiveness in aligning language model responses with human preferences\cite{RLHF, rlhf2, rlhf3, rlhf4, rlhf5}. Its three stage pipeline, including supervised fine-tuning(SFT)\cite{SFT1, SFT2, SFT3, SFT4}, reward modeling\cite{RW1, RW2, RW3, RW4}, and reinforcement learning optimization\cite{PPO, RL1}, increases the overall complexity of RLHF. DPO\cite{DPO} mitigates this limitation by constructing preference datasets offline and directly optimizing the model using its designed loss function. IPO\cite{IPO} analyze the DPO, proposed a generalized preference theory, and mitigated the overfitting issues present in the DPO. SimPO\cite{simpo} redefines implicit rewards in DPO, transforming it into a reference-free preference optimization method with remarkable effectiveness. However, it relies on extensive experimentation to determine optimal hyperparameter selections. $\beta$-DPO\cite{beta} investigates the relationship between hyperparameters and the data distribution. However, it requires fine-tuning around a preset hyperparameter value. Similarly, AlphaDPO\cite{aDPO} adopts an analogous strategy to adjust the $\gamma$ in SimPO, but it reintroduces a reference model, thereby increasing resource waste. At the same time, they also introduce a new hyperparameter, $\alpha$. SimPER\cite{simper} uses model perplexity as the reward and proposes a hyperparameter-free preference optimization method; however, its performance remains unsatisfactory. This paper proposes $\xi$-DPO, which retains the simplicity of SimPO while avoiding its cumbersome hyperparameter selection process, and achieves strong empirical performance.

	\section{Preliminaries}
	\textbf{Preference modeling.} Given a dataset $D=\{(x, y_w, y_l)\}$, it consists of prompts $x$ and paired responses ($y_w$, $y_l$), where $y_w$ is the chosen response(winning) and $y_l$ is the rejected response(losing). Christiano et al.\cite{rlhf2} propose using the Bradley-Terry model(BT model)\cite{bt} for preference modeling to optimize the reward model, $r(y, x)$:
	\begin{equation}
		\begin{aligned}
			p(y_w \succ y_l \mid x)
			&= \frac{\exp\big(r(y_w, x)\big)}{\exp\big(r(y_w, x)\big)+\exp\big(r(y_l,x)\big)}
			\\
			&= \sigma\big(r(y_w, x)-r(y_l, x)\big)
			\label{eq:bt}
		\end{aligned}
	\end{equation}
	where $\sigma$ is the sigmoid function. $y_w \succ y_l$ indicates that chosen response $y_w$ is strictly preferred over rejected response $y_l$. The purpose of preference modeling is to learn a reward model that assigns scores to the outputs of target policy $\pi_\theta$ (preference prediction).
	
	\textbf{Reinforcement Learning from Human Feedback
		(RLHF).} Based on the learned BT reward model 
	$r(x, y)$, the responses generated by the target policy $\pi_\theta$ are scored, while the policy is simultaneously constrained to remain close to the reference policy $\pi_{ref}$\cite{rlhf4, rlhf6}. Specifically, the formula is as follows:
	\begin{equation}
		\label{rlhf}
		\max_{\pi_\theta}\;
		\mathbb{E}_{x \sim \mathcal{D},\, y \sim \pi_\theta(y \mid x)}
		\left[
		r(x, y)
		\right]
		-
		\beta \mathbb{D}_{\mathrm{KL}}
		\left[
		\pi_\theta(y \mid x)\,\|\,\pi_{\mathrm{ref}}(y \mid x)
		\right]
	\end{equation}
	Typically, the reference policy is initialized to be identical to the target policy, and it is a supervised fine-tuning(SFT) model.
	
	\textbf{Direct Preference Optimization (DPO).} Following Equation \ref{rlhf}, Rafailov et al.\cite{DPO} derive an implicit form of the reward model. By connecting it to the BT model, they unify reward modeling and reinforcement learning into a single stage:
	\begin{equation}
		p(y_w \succ y_l \mid x)
		= \sigma\left(\beta \log \frac{\pi_\theta(y_w \mid x)}{\pi_\theta(y_l \mid x)} - \beta \log \frac{\pi_{\mathrm{ref}}(y_w \mid x)}{\pi_{\mathrm{ref}}(y_l \mid x)}\right)
	\end{equation}
	
	where $\beta \log \frac{\pi_\theta(y \mid x)}{\pi_{\mathrm{ref}}(y \mid x)}$ is the implicit form of $r(x,y)$. Finally, the DPO loss is defined as follows:

	\begin{equation}
		\begin{aligned}
			&
			\mathcal{L}_{\mathrm{DPO}}
			=
			-\mathbb{E}_{(x,y_w,y_l)\sim\mathcal{D}}
			\Biggl[\\
			&
			\log \sigma
			\left(
			\beta \log \frac{\pi_\theta(y_w \mid x)}{\pi_{\mathrm{ref}}(y_w \mid x)}
			-
			\beta \log \frac{\pi_\theta(y_l \mid x)}{\pi_{\mathrm{ref}}(y_l \mid x)}
			\right)
			\Biggr]
		\end{aligned}
	\end{equation}
	
	\textbf{Simple Preference Optimization (SimPO).} When DPO is applied, an additional reference model with the same optimization objective is loaded into memory, which increases resource requirements. Motivated by this observation, SimPO\cite{simpo} explores a reference-free reward mechanism. However, simply removing the reference model from DPO leads to an imbalance between the chosen and rejected responses. The main reason is that these responses often differ in length, and longer responses are naturally associated with larger numerical values. To address this issue, SimPO normalizes the responses and defines the reward as $\frac{\beta}{|y|}\log\pi_\theta(y|x)$. Based on this, the reward gap is also constrained by defining reward margin $\gamma$. Ultimately, the SimPO loss is defined as:
	\begin{equation}
		\begin{aligned}
			&
			\mathcal{L}_{\mathrm{SimPO}}
			=
			-\mathbb{E}_{(x,y_w,y_l)\sim\mathcal{D}}
			\Biggl[\\
			&
			\log \sigma
			\left(
			\frac{\beta}{|y_w|} \log \pi_\theta(y_w \mid x)
			-
			\frac{\beta}{|y_l|} \log \pi_\theta(y_l \mid x) - \gamma
			\right)
			\Biggr]
			\label{eq:simpo}
		\end{aligned}
	\end{equation}
	where $|y|$ is the length of the response of the policy $\pi_{\theta}(y \mid x)$.
	
	\section{Hyperparameter Role Analysis in SimPO}
	
	This section provides a comprehensive analysis of the roles of the hyperparameters in SimPO.

	\subsection{The Role of $\beta$}
	\label{analysis}
	Given prompt $x$ and its corresponding chosen ($y_w$) and rejected ($y_l$) responses. According to Equation \ref{eq:simpo_}, the primary optimization objective is to minimize the negative log-likelihood of the probability of the gap between the chosen and rejected rewards ($r(y_w, x)$ and $r(y_l, x)$), that is:
	\begin{equation}
		-log\sigma(\beta(r^*(y_w, x) -r^*(y_l, x)))
		\label{overview_objective}
	\end{equation}
	where $r^*(y,x)=\frac{log\pi_\theta(y|x)}{|y|}$. For now, we will set $\gamma$ aside, as it can be treated as a bias that does not affect our analysis of the distribution of reward gaps. For simplicity, we refer to the chosen reward as chosen and the rejected reward as rejected.

\begin{figure}[t]
		\centering
		
		\includegraphics[width=\linewidth]{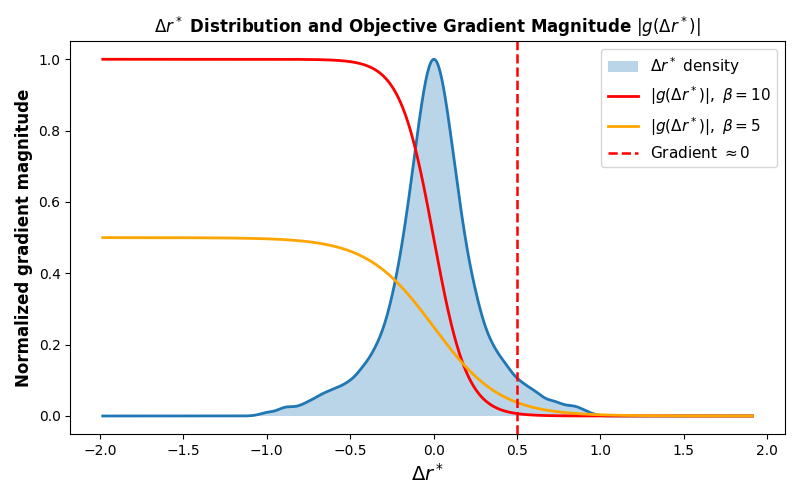}
		\caption{Distribution of $\Delta r^*$ and the objective gradient magnitude $|g(\Delta r^*)|$. A larger $\beta$ sharpens the gradient decay and filters more tail-region samples (right). Moreover, increasing $\beta$ strengthens the gradients of head samples but suppresses those of tail samples.}
		\label{fig:distribution}
		
	\end{figure}

	For a given dataset, the distribution of reward gap ($\Delta r^*=r^*(y_w, x) -r^*(y_l, x)$) will exhibit the following situations: 1). Chosen is far greater than rejected ($\Delta r^* \gg 0$); 2). Chosen is far less than rejected ($\Delta r^* \ll 0$); 3). Chosen is close to rejected. As shown in Figure \ref{fig:distribution}, the first two cases are referred to as high reward gap (the tail and head of the distribution, respectively), and the third case is low-gap(the middle of the distribution). Hereafter, we use the terms tail and head regions to denote the first two cases. The samples that fall into the third case are considered as normal and help maintain training stability, since it is sufficient to enforce a certain constraint such that the chosen remains larger than the rejected. The samples in the tail regions exhibit strong reward signals and can facilitate the optimization process when the hyperparameters are properly selected; otherwise, an originally large reward gap may be undesirably reduced. The most problematic samples are those in the head regions, which could severely disrupt the training process. 
	
	{Previous studies\cite{DPO, simpo, aDPO} generally treat $\beta$ as a scaling factor. We find that this characterization is incomplete, because $\beta$ does not scale the gradients uniformly across the reward-gap distribution. Specifically, the gradient magnitude $|g(\Delta r^*)|$ is positively associated with $\beta$ in the head region (left region of the distribution), but negatively associated with $\beta$ in the tail region. As illustrated in Figure \ref{fig:distribution}. increasing $\beta$ shifts forward the point at which the gradient approaches zero, causing more high gap samples in the tail region to receive negligible gradient updates. For example, at $\Delta r^*=0.5$, the gradient magnitude is already close to zero when $\beta=10$, whereas it is remains activated when $\beta=5$. Therefore, in addition to controlling the gradient scale, $\beta$ implicitly determines how strongly high-gap tail samples are filtered during optimization. Since changing $\beta$ induces coupled and opposite impact across the reward-gap distribution: the gradients of both head- and tail-region samples are affected simultaneously, but with different magnitudes and in opposite directions. Consequently, one region cannot be adjusted independently without altering the other. This entangled and non-uniform effect makes the optimization outcome highly sensitive to β and complicates the selection of a robust value.}

	
	
	This insight also explains the empirical observation of $\beta$-DPO\cite{beta}, where a smaller $\beta$ is preferred for low-gap data and a larger one for high-gap data. For low-gap data, samples are concentrated around the middle region, and a smaller $\beta$ maintains a smoother gradient profile, allowing these fine-grained samples to contribute more uniformly. For high-gap data, a larger $\beta$ strengthens the optimization of head-region samples where the chosen and rejected responses are incorrectly ordered, while suppressing tail-region samples that are already well separated. However, the relative importance of the head and tail regions varies across datasets, making the effect of changing $\beta$ inherently uncertain. Together with the symmetric coupling between head amplification and tail suppression, this oppositely directed symmetry makes $\beta$ highly sensitive to the reward-gap distribution and difficult to tune robustly.

	
	
	
	\subsection{The Role of $\gamma$}
	\label{sec:gamma}

	We analyze the effect of $\gamma$ from two complementary perspectives: token-level investigation and comparative analysis between SimPO and DPO.
	
	\textbf{Token-level Investigation.} For qualitative analysis, given a prompt $x$, we assume that the chosen and rejected responses have the same length, denoted by $|y|$. Then, the optimization objective of SimPO can be represented by:
	
	\begin{equation}
		\begin{aligned}
			&\frac{\beta}{|y|}\log \pi_{\theta}(y_w \mid x)-\frac{\beta}{|y|}\log \pi_{\theta}(y_l \mid x)-\gamma \\
			=&\ \frac{\beta}{|y|}\sum_{i=1}^{n}\log \pi_{\theta}(y_w^i \mid y_w^{i-1}, \ldots, x)\\
			&\quad-\frac{\beta}{|y|}\sum_{i=1}^{n}\log \pi_{\theta}(y_l^i \mid y_l^{i-1}, \ldots, x)-\gamma \\
			=&\ \Bigl[
			\left(\frac{\beta}{|y|}\log \pi_{\theta}(y_w^1 \mid x)-\frac{\beta}{|y|}\log \pi_{\theta}(y_l^1 \mid x)-\gamma_1\right) \\
			&
			+\left(\frac{\beta}{|y|}\log \pi_{\theta}(y_w^2 \mid y_w^1, x)-\frac{\beta}{|y|}\log \pi_{\theta}(y_l^2 \mid y_l^1, x)-\gamma_2\right)+\cdots
			\Bigr]
		\end{aligned}
		\label{eq:expand}
	\end{equation}
	This suggests that $\gamma$ can be viewed as imposing a token-level constraint. For low-gap pairs, the chosen and rejected responses typically exhibit substantial token overlap, implying that these shared or highly similar tokens require little additional separation($\gamma_i=0$). As a result, the cumulative constraint induced by $\gamma$ should be smaller. This also explains why low-gap datasets favor a smaller $\gamma$: the purpose is not to enforce an unnecessarily large margin between chosen and rejected responses, but to adapt the constraint strength to the intrinsic characteristics of data.
	
	\textbf{Comparative Analysis between DPO and SimPO.} Junkang et al.\cite{aDPO} have reached the conclusion that $\gamma$ is actually an implicit representation of the reference policy. Here, we leverage their insights to support our hypothesis. This is illustrated by the following equation:
	\begin{align}
			&\mathcal{L}_{\mathrm{DPO}}
			=\ -\mathbb{E}_{(x,y_w,y_l)\sim\mathcal{D}}
			\Biggl[\notag\\*
			&
			\log \sigma
			\left(
			\frac{\beta}{|y_w|}\log \frac{\pi_{\theta}(y_w\mid x)}{\pi_{\mathrm{ref}}(y_w\mid x)}
			-
			\frac{\beta}{|y_l|}\log \frac{\pi_{\theta}(y_l\mid x)}{\pi_{\mathrm{ref}}(y_l\mid x)}
			\right)
			\Biggr] \notag\displaybreak[1]\\
			&=\ -\mathbb{E}_{(x,y_w,y_l)\sim\mathcal{D}}
			\Biggl[\notag\\*
			&
			\log \sigma
			\Biggl(
			\frac{\beta}{|y_w|}\log \pi_{\theta}(y_w\mid x)
			-
			\frac{\beta}{|y_l|}\log \pi_{\theta}(y_l\mid x)
			\notag\\*
			&\quad
			-
			\Bigl(
			\frac{\beta}{|y_w|}\log \pi_{\mathrm{ref}}(y_w\mid x)
			-
			\frac{\beta}{|y_l|}\log \pi_{\mathrm{ref}}(y_l\mid x)
			\Bigr)
			\Biggr)
			\Biggr] \notag\displaybreak[1]\\
			&=\ -\mathbb{E}_{(x,y_w,y_l)\sim\mathcal{D}}
			\Biggl[\notag\\*
			&
			\log \sigma
			\left(
			\frac{\beta}{|y_w|}\log \pi_{\theta}(y_w\mid x)
			-
			\frac{\beta}{|y_l|}\log \pi_{\theta}(y_l\mid x)
			-\gamma
			\right)
			\Biggr] \notag\\*
			&=\ \mathcal{L}_{\mathrm{SimPO}}
	\end{align}
	As can be seen from the above equation, $\gamma=\frac{\beta}{|y_w|}\log \pi_{\mathrm{ref}}(y_w\mid x)
	-
	\frac{\beta}{|y_l|}\log \pi_{\mathrm{ref}}(y_l\mid x)$. This means that a low-gap dataset naturally requires a smaller $\gamma$; conversely, using a large $\gamma$ would deviate from the actual performance of the reference policy. This indicates a potential mismatch between the behavior of $\gamma$ and its intended role of artificially widening the reward gap.
	
	\section{Our Methodology}
	
	Based on our analysis of $\beta$ and $\gamma$, we present our $\xi$-DPO solution with two objectives: to eliminate the high sensitivity of $\beta$, where even small changes can substantially affect the log-sigmoid gradient, and to introduce a new, quantifiable margin that enforces a stronger constraint on the reward gap between chosen and rejected responses.
	
	\subsection{Definition of $\xi$-DPO}
	\label{sec:xi_definition}
	
	First, we apply a logit transformation to the reward objective in the Equation \ref{overview_objective} to mitigate the subtle influence of the log-sigmoid gradient on the optimization process, which is as follows:
	\begin{equation}
		f(x)=\log\left(\frac{x}{1-x}\right)
		\label{eq:new_objective}
	\end{equation}
	
	After applying the transformation to the original objective as shown in Equation \ref{overview_objective}, the new objective now becomes $\beta(r^*(y_w, x) -r^*(y_l, x))$. Based on the analysis in IPO\cite{IPO}, such a transformation does not alter the optimal objective. Accordingly, the objective of SimPO\cite{simpo} can be represented by:
	\begin{equation}
		\frac{\beta}{|y_w|} \log \pi_\theta(y_w \mid x)
		-
		\frac{\beta}{|y_l|} \log \pi_\theta(y_l \mid x) - \gamma
		\label{eq:simpo_new_objective}
	\end{equation}
	
	The current objective of SimPO has shifted from maximizing the likelihood of the reward gap to minimizing the distance between the reward gap and $\gamma$. Here, the $\gamma$ is considered the optimal gap between chosen and rejected, and the final objective can be obtained using the least-squares method:
	\begin{equation}
		(\frac{\beta}{|y_w|} \log \pi_\theta(y_w \mid x)
		-
		\frac{\beta}{|y_l|} \log \pi_\theta(y_l \mid x) - \gamma)^2
		\label{eq:simpo_final_objective}
	\end{equation}
	Then, we redefine the reward through normalization to ensure that it is bounded: 
	\begin{equation}
		\frac{r_*}{|r_w+r_l|}
		\label{redefine}
	\end{equation}
	where $r_*$ is chosen($r_w$) or rejected($r_l$), and the chosen and rejected rewards constitute a reward ratio. 
	
	Accordingly, we introduce a ratio reward margin $\xi$ to control the gap between ratio rewards. For simplicity, we refer to it as the ratio margin hereafter. By normalizing the reward $\frac{\beta}{|y_*|} \log \pi_\theta(y_*|x)$ in SimPO according to Equation \ref{redefine}, the objective can be defined as follows:
	\begin{equation}
		\left(
		\frac{logp_w -logp_l}
		{\left|logp_w + logp_l\right|}
		-\xi
		\right)^2
		\label{eq:eqatio}
	\end{equation}
	where $logp_*$ is  $\frac{\log \pi_\theta(y_*|x)}{|y_*|}$. It is noteworthy that the reward definition in Equation \ref{redefine} offers two advantages: 1) it ensures that the reward remains bounded while constraining the ratio margin, $\xi\in [-1, 1]$; 2) it eliminates the dependence on $\beta$ and defines $\xi$ on a bounded ratio reward space, making the margin more interpretable and less sensitive, while allowing it to be selected from the initial reward gap distribution rather than through joint tuning with $\beta$.
	
	\noindent\textbf{Proof of order preservation.} {For $\beta>0$, let $r_*=\beta a_*$, where $a_*=|y_*|^{-1}\log\pi_\theta(y_*\mid x)$ for $*\in\{w,l\}$. Write $d=a_w-a_l$. Since log probabilities are non-positive, $a_w\leq0$ and $a_l\leq0$. Assume $|a_w+a_l|>0$ and define $s=\operatorname{sg}(|a_w+a_l|)$, where $\operatorname{sg}$ is stop-gradient, implemented with \texttt{detach()}. In the forward pass, this operation leaves the denominator unchanged. The normalized gap therefore satisfies}
	\begin{equation}
	    \begin{aligned}
	        m&=\frac{r_w-r_l}{\operatorname{sg}(|r_w+r_l|)}
	          ={\frac{\beta(a_w-a_l)}{\beta s}}=\frac{d}{s},\\
	        \operatorname{sign}(m)&=\operatorname{sign}(d),\qquad
	        \nabla_\theta m=\frac{1}{s}\nabla_\theta d.
	    \end{aligned}
	    \label{eq:ratio_order}
	\end{equation}
	{Because $s>0$, dividing by $s$ cannot reverse the sign of $d$: $m>0$ if and only if $a_w>a_l$, and ties remain ties. Thus, normalization preserves the chosen--rejected ordering within each pair while cancelling the common factor $\beta$. Moreover, $|d|\leq|a_w|+|a_l|=|a_w+a_l|=s$, so $|m|\leq1$, which establishes the bounded ratio reward space.}
	
	{During backpropagation, the denominator is treated as constant, so the chain rule gives the gradient identity above. The per-sample reward-gap gradient direction is therefore preserved, with its magnitude rescaled by the positive factor $1/s$. This statement concerns individual reward gaps; the sample-dependent rescaling can still change their relative contributions to a minibatch update.}
	
	
	However, the objective as shown in Equation \ref{eq:eqatio} still has an undesirable problem. Specifically, if the chosen is larger than rejected, smaller values of $\xi$ will cause the chosen to decrease and the rejected to increase during optimization, thereby narrowing the originally large gap between the two back to $\xi$. Such samples are concentrated at the tail end of distribution as shown in Figure \ref{fig:distribution}. These tail samples do not need to be included in model training, as the model already performs very well on these samples. Therefore, we define the final $\xi$-DPO loss as:
	\begin{equation}
		\begin{aligned}
			&
			\min_{\theta}\;
			\mathbb{E}_{(x,y_w,y_l)\sim\mathcal{D}}
			\Biggl[\\
			&
			\operatorname{LeakyReLU}
			\left(
			\xi
			-
			\left(\frac{
				\frac{1}{|y_w|}\log \pi_{\theta}(y_w \mid x)
				-
				\frac{1}{|y_l|}\log \pi_{\theta}(y_l \mid x)
			}{
				\left|
				\frac{1}{|y_w|}\log \pi_{\theta}(y_w \mid x)
				+
				\frac{1}{|y_l|}\log \pi_{\theta}(y_l \mid x)
				\right|
			}\right)
			\right)^2
			\Biggr]
			\label{eq:our_loss}
		\end{aligned}
	\end{equation}
	where $\operatorname{LeakyReLU}$ causes the gradients of samples whose reward gaps exceed $\xi$ to approach zero. It is important to note that the normalization factor $		\left|
	\frac{1}{|y_w|}\log \pi_{\theta}(y_w \mid x)
	+
	\frac{1}{|y_l|}\log \pi_{\theta}(y_l \mid x)
	\right|$ does not participate in gradient updates. It is treated only as a constant coefficient used to normalize the reward into a ratio form. A visual illustration of the LeakyReLU behavior described above, together with additional details, is provided in the Appendix \ref{sec:worelu}.
	
	
	\subsection{Setting of $\xi$}
	\label{sec:xi}

	
	The key factor underlying this desirable property of $\xi$-DPO is $\xi$. The choice of $\xi$ mainly depends on the extent to which the gap between chosen and rejected should be enlarged. Meanwhile, when combined with $\operatorname{LeakyReLU}$, it can effectively filter out tail samples, thereby controlling the strength of the reward signal. Specifically, filtering out more tail samples reduces the strength of the reward signal, while retaining more tail samples increases it. This behavior varies across datasets, since in low-gap datasets the distribution of gaps is typically more concentrated, with relatively few samples located at the head and tail regions shown in Figure \ref{fig:distribution}, resulting in a generally moderate reward gap. In contrast, high-gap datasets exhibit the opposite pattern. Therefore, $\xi$ should be chosen such that it exceeds the reward gap of the vast majority of samples, while also controlling whether more strong reward signals (tail samples) should be introduced for datasets of different quality. We define $\xi$ as follows:
	
	\begin{equation}
		\xi = Q_{t}\!\left(\{m_i\}^N_{i=1}\right)
		\label{eq:quntile}
	\end{equation}
	where $Q_{t}(\cdot)$ denotes the $t$-th quantile of the distribution $\{m_i\}^N_{i=1}$, $N$ is the number of samples. $m_i$ denotes the reward gap in the current sample, which is computed using the initial model, $\pi(y|x)$. For $\xi$-DPO,  $m_i$ is:
	
	\begin{equation}
		\frac{
			\frac{1}{|y_w|}\log \pi(y_w^i \mid x^i)
			-
			\frac{1}{|y_l|}\log \pi(y_l^i \mid x^i)
		}{
			\left|
			\frac{1}{|y_w|}\log \pi(y_w^i \mid x^i)
			+
			\frac{1}{|y_l|}\log \pi(y_l^i \mid x^i)
			\right|
		}
	\end{equation}
	
	
	{Under the $\xi$-DPO formulation, sample filtering is no longer governed by the log-sigmoid slope induced by $\beta$, but is adaptively controlled by the ratio margin $\xi$. The samples with ratio reward gaps below $\xi$ remain active, while those already exceeding $\xi$ are down weighted through LeakyReLU. Thus, $\xi$ acts as both a semantic margin and an automatic filtering threshold. Specifically, for high-gap datasets we use a larger $\xi$ because their reward gaps are generally larger, and a small margin may cause optimization to stop too early. Moreover, retaining tail-region samples with strong positive reward signals can help counteract head-region samples where rejected responses dominate, encouraging the reward gap to shift from negative to positive. For low-gap datasets, we impose a milder reward strength, since overly strong reward signals may distort the model’s original distribution and increase the risk of overfitting. Accordingly, we fix $t=95\%$ for low-gap datasets and $t=99.9\%$ for high-gap datasets. We have conducted sensitivity analysis in Section \ref{sec:main_res}, showing that the performance remains stable when $t$ varies within 90\%--95\% for low-gap data and 97\%--99.9\% for high-gap data. This indicates that $\xi$ is robust to moderate variations in the quantile level. A more detailed analysis of $\xi$ is provided in Appendix \ref{sec:explaination}.}
	
	{It is noteworthy that based on our previous analysis, the tuning difficulty of SimPO arises not merely from having two hyperparameters, but from the fact that $\beta$ and $\gamma$ are difficult-to-quantify, and have coupled  effect. Specifically, $\beta$ changes the slope and saturation region of the sigmoid, thereby implicitly controlling which low-gap and high-gap samples contribute, while $\gamma$ determines the required separation between the chosen and rejected responses relative to the dataset’s intrinsic reward-gap structure. Consequently, changing $\beta$ also changes the effective strength of a fixed $\gamma$, and the same ($\beta$, $\gamma$) configuration can behave very differently across datasets. Neither parameter can be directly derived from observable statistics of the preference data, so SimPO typically requires a two-dimensional grid search over multiple $\beta$ and $\gamma$ combinations.
		In contrast, ξ-DPO uses a fixed category-level rule, which can be directly computed from the initial reward-gap distribution, to replace the $\beta$ and $\gamma$. $\xi$ is much easier to set, because the gap type of the data does not require posterior judgment. 
		In practice, the gap type can be determined from the reward model used to construct the preference dataset. Strong reward models have fine-grained ranking capabilities, enabling them to reliably distinguish responses of comparable quality and thus construct low-gap preference data. Conversely, weaker reward models tend to distinguish responses only when their quality differences are more evident, resulting in high-gap preference data.}

	
	
	
	\section{Experiments}
	
	\begin{table*}[!t]
		\centering
		\caption{Comparison of different preference optimization methods on AlpacaEval 2 and MT-Bench across four SFT models.}
		\label{tab:main_results}
		\scriptsize
		\setlength{\tabcolsep}{3pt}
		\renewcommand{\arraystretch}{1.02}
		\resizebox{\textwidth}{!}{%
			\begin{tabular}{lcccccccccccc}
				\specialrule{1.1pt}{0pt}{0pt}
				\multirow{3}{*}{\textbf{Method}}
				& \multicolumn{3}{c}{\textbf{Llama3-Instruct (8B)}}
				& \multicolumn{3}{c}{\textbf{Mistral-Instruct (7B)}}
				& \multicolumn{3}{c}{\textbf{Llama3-Instruct v0.2 (8B)}}
				& \multicolumn{3}{c}{\textbf{Gemma2-Instruct (9B)}} \\
				\cmidrule(lr){2-4} \cmidrule(lr){5-7} \cmidrule(lr){8-10} \cmidrule(lr){11-13}
				& \multicolumn{2}{c}{\textbf{AlpacaEval 2}} & \textbf{MT-Bench}
				& \multicolumn{2}{c}{\textbf{AlpacaEval 2}} & \textbf{MT-Bench}
				& \multicolumn{2}{c}{\textbf{AlpacaEval 2}} & \textbf{MT-Bench}
				& \multicolumn{2}{c}{\textbf{AlpacaEval 2}} & \textbf{MT-Bench} \\
				\cmidrule(lr){2-3} \cmidrule(lr){4-4}
				\cmidrule(lr){5-6} \cmidrule(lr){7-7}
				\cmidrule(lr){8-9} \cmidrule(lr){10-10}
				\cmidrule(lr){11-12} \cmidrule(lr){13-13}
				& \textbf{LC (\%)} & \textbf{WR (\%)} & \textbf{GPT-4}
				& \textbf{LC (\%)} & \textbf{WR (\%)} & \textbf{GPT-4}
				& \textbf{LC (\%)} & \textbf{WR (\%)} & \textbf{GPT-4}
				& \textbf{LC (\%)} & \textbf{WR (\%)} & \textbf{GPT-4} \\
				\midrule
				SFT
				& 24.0 & 23.6 & 8.1
				& 19.0 & 15.4 & 7.5
				& 24.0 & 23.6 & 8.1
				& 48.7 & 36.5 & 8.5 \\
				\midrule
				DPO
				& 40.2 & 38.1 & 8.0
				& 20.3 & 17.9 & 7.6
				& 51.9 & 50.8 & 8.2
				& 70.4 & \underline{66.9} & 8.5 \\
				IPO
				& 35.9 & 34.4 & \textbf{8.3}
				& 22.3 & 18.6 & \textbf{7.8}
				& 40.6 & 39.6 & 8.2
				& 62.6 & 58.4 & - \\
				CPO
				& 29.6 & 34.4 & 8.0
				& 26.2 & 31.7 & 7.5
				& 36.5 & 40.8 & 8.2
				& 56.4 & 53.4 & - \\
				KTO
				& 38.3 & 34.1 & \underline{8.2}
				& 19.4 & 20.3 & \underline{7.7}
				& 41.4 & 36.4 & 8.2
				& 61.7 & 55.5 & - \\
				ORPO
				& 31.6 & 29.8 & 8.0
				& 24.0 & 23.0 & \underline{7.7}
				& 36.5 & 33.1 & \textbf{8.3}
				& 56.2 & 46.7 & - \\
				R-DPO
				& 40.3 & 37.3 & 8.0
				& 21.4 & 22.2 & 7.5
				& 51.6 & \underline{50.7} & 8.2
				& 68.3 & \underline{66.9} & - \\
				SimPO
				& 43.8 & 38.0 & 8.0
				& 30.2 & 32.1 & 7.6
				& 55.6 & 49.6 & 8.0
				& 72.4 & 65.0 & - \\
				AlphaDPO
				& \underline{46.6} & \underline{38.1} & -
				& \underline{32.3} & \underline{32.6} & -
				& \textbf{58.7} & \textbf{51.1} & -
				& 73.4 & 66.1 & - \\
				\midrule
				$\xi$-DPO
				& \textbf{47.1} & \textbf{39.1} & \underline{8.2}
				& \textbf{33.2} & \textbf{36.5} & \underline{7.7}
				& \underline{57.5} & 50.5 & 8.0
				& \textbf{75.4} & \textbf{67.5} & \textbf{9.0} \\
				\specialrule{1.1pt}{0pt}{0pt}
			\end{tabular}%
		}
		
	\end{table*}
	
	\begin{table*}[!t]
		\captionsetup{skip=4pt}
		\centering
		\caption{\textbf{Sensitivity Analysis of $\xi$-DPO}. We report the sensitivity of $\xi$, computed based on different quantiles, by varying the quantile threshold on both a low-gap dataset constructed with a stronger reward model and a high-gap dataset constructed in the opposite setting.}
		\label{tab:ablation}
		
		\begin{minipage}{0.49\linewidth}
			\centering
			\captionof*{table}{(a) Llama3-Instruct-v0.2 (Low-gap dataset)}
			\resizebox{\linewidth}{!}{
				\begin{tabular}{lcccccc}
					\toprule
					\multirow{2}{*}{Method} 
					& \multicolumn{2}{c}{$t=0.90$} 
					& \multicolumn{2}{c}{$t=0.92$} 
					& \multicolumn{2}{c}{$t=0.95$} \\
					\cmidrule(lr){2-3} \cmidrule(lr){4-5} \cmidrule(lr){6-7}
					& LC & WR & LC & WR & LC & WR \\
					\midrule
					$\xi$-DPO 
					& 56.4 & 49.1 & 57.2 & 49.9 & \textbf{57.5} & \textbf{50.5} \\
					\bottomrule
				\end{tabular}
			}
		\end{minipage}
		\hfill
		\begin{minipage}{0.49\linewidth}
			\centering
			\captionof*{table}{(b) Mistral-Instruct (High-gap dataset)}
			\resizebox{\linewidth}{!}{
				\begin{tabular}{lcccccc}
					\toprule
					\multirow{2}{*}{Method} 
					& \multicolumn{2}{c}{$t=0.97$} 
					& \multicolumn{2}{c}{$t=0.98$} 
					& \multicolumn{2}{c}{$t=0.999$} \\
					\cmidrule(lr){2-3} \cmidrule(lr){4-5} \cmidrule(lr){6-7}
					& LC & WR & LC & WR & LC & WR \\
					\midrule
					$\xi$-DPO 
					& 32.4 & 35.9 & 32.9 & 36.1 & \textbf{33.2} & \textbf{36.5} \\
					\bottomrule
				\end{tabular}
			}
		\end{minipage}
		
	\end{table*}
	
	\subsection{Experimental Setup}
	
	\textbf{Datasets and models.} Following Meng et al.\cite{simpo} and Junkang et al.\cite{aDPO}, we use 4 datasets: 1) \href{https://huggingface.co/datasets/princeton-nlp/llama3-ultrafeedback}{princeton-nlp/llama3-ultrafeedback}; 2) \href{https://huggingface.co/datasets/princeton-nlp/llama3-ultrafeedback-armorm}{princeton-nlp/llama3-ultrafeedback-armorm}; 3) \href{https://huggingface.co/datasets/princeton-nlp/mistral-instruct-ultrafeedback}{princeton-nlp/mistral-instruct-ultrafeedback}; 4) \href{https://huggingface.co/datasets/princeton-nlp/gemma2-ultrafeedback-armorm}{princeton-nlp/gemma2-ultrafeedback-armorm} for \href{https://huggingface.co/meta-llama/Meta-Llama-3-8B-Instruct}{Llama3-8B-Instruct}\cite{llama3}, \href{https://huggingface.co/mistralai/Mistral-7B-Instruct-v0.2}{Mistral-7B-Instruct}\cite{mistral} and \href{https://huggingface.co/google/gemma-2-9b-it}{Gemma2-9B-Instruct}\cite{gemma} training. Llama3-8B-Instruct is trained separately on datasets 1 and 2. The key difference is that dataset 2 is constructed using a stronger reward model \href{https://huggingface.co/RLHFlow/ArmoRM-Llama3-8B-v0.1}{RLHFlow/ArmoRM-Llama3-8B-v0.1}\cite{strong}. We denote the trained model as Llama3-Instruct-v0.2. 
	
	\textbf{$\xi$ setting.} Among the four datasets described above, the datasets 2 and 4, which are constructed using a stronger reward model\allowbreak(ArmoRM), are considered to be low-gap datasets. Using Equation \ref{eq:quntile}, we compute the values of $\xi$ for these two datasets as 0.35 and 0.28, respectively, which correspond to the 95th quantile of the reward gap distribution. In contrast, for the datasets 1 and 3, $\xi$ is set to 0.9 and 0.45 respectively, corresponding to the 99.9th percentile. We provide more explanations on the choice of the 95th or 99.9th quantiles in Appendix \ref{sec:explaination}.
	
	\subsection{Benchmarks and Compared Methods}
	
	\textbf{Benchmarks.} AlpacaEval 2\cite{alpacal} and MT-Bench\cite{mmt} are widely used benchmarks for evaluating the response quality and instruction-following ability of large language models. AlpacaEval 2 typically uses a judge model to compare the responses of a target model against those of a baseline model and reports metrics such as win rate(WR) and length-controlled win rate(LC), whereas MT-Bench usually evaluates a model’s responses to multi-turn questions using a strong LLM judge and reports an overall score. Following the setups adopted in AlphaDPO\cite{aDPO} and SimPO\cite{simpo}. GPT-4 Turbo is used as both the baseline model and the judge model for AlpacaEval 2, whereas MT-Bench uses GPT-4 to evaluate the target model’s outputs. {We additionally evaluate on the verifiable Text-to-SQL task to further support the reliability of the proposed ξ-selection rule, with additional details provided in Appendix \ref{sec:sql}.}
	
	\textbf{Compared Methods.} We compare our proposed $\xi$-DPO with several state-of-the-art preference optimization algorithms: DPO\cite{DPO}, IPO\cite{IPO}, CPO\cite{COT}, KTO\cite{KTO}, ORPO\cite{ORPO}, R-DPO\cite{R-DPO}, SimPO\cite{simpo}, AlphaDPO\cite{aDPO}.
	
	\subsection{Main results and Sensitivity analysis}
	\label{sec:main_res}
	
	\textbf{Comparative Evaluation Results.} As shown in Table \ref{tab:main_results}, $\xi$-DPO achieves the best or highly competitive performance across all the datasets, outperforming existing alternatives on AlpacaEval 2 and remaining competitive on MT-Bench. Especially, $\xi$-DPO achieves a 12.0\% relative improvement in win rate over AlphaDPO on Mistral-Instruct. It also consistently outperforms the best SimPO performance achieved after multiple rounds of careful hyperparameter tuning. 
	
	The only exception is on Llama3-Instruct v0.2, where $\xi$-DPO slightly underperforms AlphaDPO. This is mainly due to the fact that the performance gain on this low-gap data setting becomes increasingly saturated, where AlphaDPO’s dynamic margin adjustment offers more fine-grained control. However, their performance gap is small, and the performance of $\xi$-DPO remains highly competitive. It is also worth pointing out that the initial hyperparameters for AlphaDPO are the same as those for SimPO, so the initial settings for its hyperparameters are also very sensitive. Furthermore, AlphaDPO introduces an additional hyperparameter $\alpha$ and a reference model $\pi_{ref}$, thereby increasing the complexity of the optimization process. These results clearly demonstrate that without relying on extensive trial-and-error hyperparameter tuning, $\xi$-DPO achieves strong performance in a more stable and efficient manner than the existing alternatives.

	
	\textbf{Sensitivity Analysis} Based on Equation \ref{eq:quntile} and our discussion of the optimal quantile, we evaluate the sensitivity of $\xi$-DPO on both low-gap and high-gap datasets. We conducted this evaluation within a reasonable range around the optimal quantiles for the low-gap and high-gap datasets. This reasonable range is related to the formulation of $\xi$-DPO, where $t$ directly controls the number of filtered samples. Therefore, retaining more than 90\% of the samples is a reasonable choice, since excessive filtering would inevitably lead to performance degradation. For the low-gap dataset, we shift $t$ from 95\% to 90\%, while for the high-gap dataset, we shift $t$ from 99.9\% to 97\%. The results in Table \ref{tab:ablation} show that this adjustment does not cause substantial performance loss, indicating that $\xi$-DPO exhibits relatively low sensitivity to this parameter. Although changing $t$ from 95\% to 90\% may appear to be a minor adjustment, the resulting values of $\xi$ differ substantially. Please refer to Table \ref{tab:qutile} in the Appendix for details, where we also provide a more detailed analysis.

	\section{Ablation Study}
	\label{sec:na}
	
	{
		W.r.t the design of $\xi$-DPO, we can observe that logit transformation, ratio normalization and the margin $\xi$ are integral to the definition of $\xi$-DPO rather than independently switchable components: logit transformation is an analytical step used to derive the margin-regression formulation, the normalization defines the ratio reward and makes $\xi$ a bounded target margin, and similarly, removing $\xi$ would eliminate the margin constraint itself and produce a different optimization objective. Therefore, on/off ablations of these components would not constitute meaningful variants of $\xi$-DPO. The only independently controllable element of the final objective is LeakyReLU.
		In this section, we conduct a corresponding ablation study on LeakyReLU, which mainly filters out samples whose reward gap has already exceeded $\xi$. We compare the effect of using LeakyReLU with that of removing it, and further include the stricter ReLU as an additional baseline. }

		
		{The evaluation results are shown in Table \ref{tab:leakyab}. They clearly demonstrate that LeakyReLU plays an important role. When the reward gap is larger than $\xi$, its gradient becomes close to zero. Moreover, when we replace LeakyReLU with ReLU, whose gradient is exactly zero in this region, the performance remains largely unchanged. This indicates that LeakyReLU not only functions similarly to ReLU by filtering samples, but also provides a less restrictive mechanism that can improve generalization to some extent. In contrast, without LeakyReLU, minimizing the squared margin error pulls reward gaps already exceeding $\xi$ back toward the target, narrowing the chosen--rejected separation. The reward curves on Llama3 presented in Appendix~\ref{sec:worelu}, which are omitted here due to space limit, also illustrate this behavior: chosen rewards decrease while rejected rewards increase. These observations complement the quantitative results in Table~\ref{tab:leakyab}.}

		\begingroup
		\setlength{\intextsep}{4pt}
		\begin{table}[H]
			\captionsetup{skip=4pt}
			\caption{\textbf{Ablation study on LeakyReLU.} We conduct an ablation study on LeakyReLU using Llama3-v0.2, a core component of $\xi$-DPO, by comparing model performance when LeakyReLU is removed and when it is replaced with ReLU.}
			\label{tab:leakyab}
			\centering
			\begin{tabular}{llc}
				\hline
				\multirow{2}{*}{Test component} & \multicolumn{2}{l}{AlpacaEval 2}                    \\ \cline{2-3} 
				& \multicolumn{1}{c}{LC}   & WR                       \\ \hline
				with LeakyReLU                  & \textbf{57.5}                     & \multicolumn{1}{c}{\textbf{50.5}} \\ \hline
				w/o LeakyReLU                   & \multicolumn{1}{c}{21.7} & 18.5                    \\ \hline
				ReLU                            & \multicolumn{1}{c}{57.2} & 50.3                    \\ \hline
			\end{tabular}
		\end{table}
		\endgroup

		\FloatBarrier
		\section{Conclusion}

		This paper investigates the semantic role of hyperparameters in simple reference-free preference optimization method (SimPO) and analyzes why the hyperparameters $\beta$ and $\gamma$ in SimPO are difficult to set. Based on the conclusions drawn from our analysis, we propose direct preference optimization via ratio reward margin, $\xi$-DPO. $\xi$-DPO has a simple objective formulation and involves only a single hyperparameter, which can be determined by computing a quantile of the prior reward-gap distribution, without requiring heuristic tuning. It effectively mitigates challenges such as the difficulty of hyperparameter selection in SimPO and reduces hyperparameter sensitivity. Our extensive experiments also demonstrate the superior performance of $\xi$-DPO.
		
		For future work, we observe that dynamically adjusting the $\xi$ threshold for each training example can potentially further improve the performance of $\xi$-DPO. It is worthy of future investigation.

		\FloatBarrier
		\bibliographystyle{plain}
		\bibliography{reference}
		
		\medskip

		\clearpage
		\appendix

		\section{Analysis of $\xi$}
		\label{sec:explaination}
\begin{figure*}[t]
			\centering
			\begin{subfigure}[b]{0.48\textwidth}
				\centering
				\includegraphics[width=\textwidth]{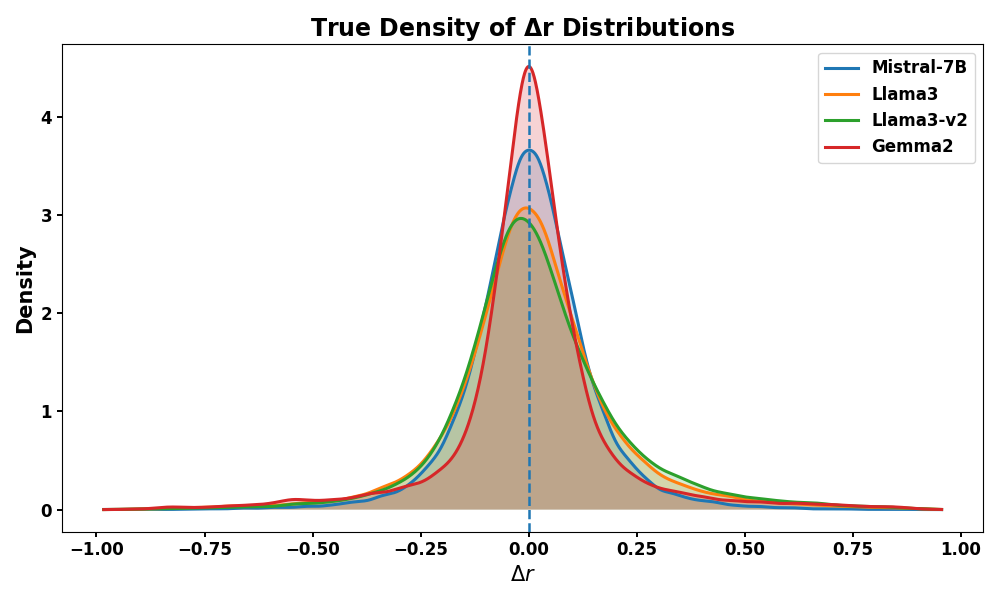}
				\caption{Density of $\Delta r$}
				\label{fig:sub1}
			\end{subfigure}
			\hfill
			\begin{subfigure}[b]{0.48\textwidth}
				\centering
				\includegraphics[width=\textwidth]{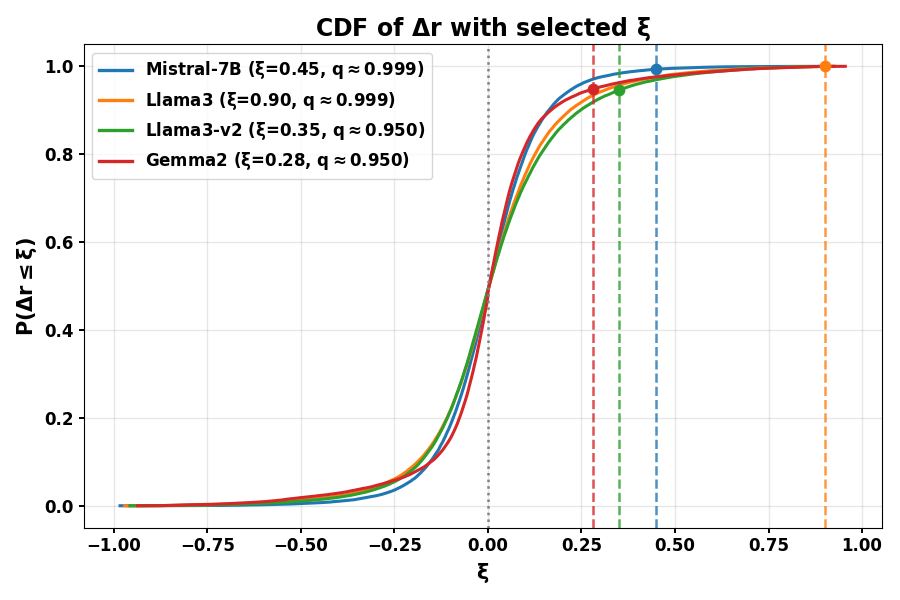}
				\caption{CDF of $\Delta r$}
				\label{fig:sub2}
			\end{subfigure}
			\caption{Density and cumulative distribution functions(CDF) of reward gaps($\Delta r$) across different datasets. (a) shows the distribution of reward gaps across different datasets, indicating that the density of reward gaps varies across datasets; the more concentrated the distribution, the smaller the ratio margin. (b) illustrates the coverage range of the datasets under different ratio margin $\xi$ settings. Notably, when the coverage for both Mistral and Llama3 is set to 99.9\%, the corresponding $\xi$ values differ, indicating that the Llama3 dataset has a more dispersed distribution.}
			\label{fig:two_figures}
		\end{figure*}

		Preference datasets typically rank responses sampled from the same policy using a reward model. Figure~\ref{fig:two_figures} illustrates the approximately symmetric reward-gap distributions and their different spreads. A rejected response may initially have higher policy probability than its chosen counterpart: correcting these head-region samples requires increasing the chosen--rejected gap. Conversely, tail-region samples already have large positive gaps. The margin $\xi$ controls their contribution: LeakyReLU downweights samples whose ratio gaps exceed $\xi$.
		
		For low-gap data with reliable rankings, a milder margin covers informative samples while filtering more well-separated tail samples. High-gap data require a stronger margin: retaining positive tail signals helps compensate for weak or incorrectly ordered head signals. Accordingly, we use $t=95\%$ and $t=99.9\%$, respectively, in Equation~\eqref{eq:quntile}. These quantiles are computed from the initial policy, without repeated model training.
		
		Table~\ref{tab:qutile} reports the full quantile statistics. The large separation between the 99th percentile and maximum motivates avoiding $t=100\%$, which can overconstrain ordinary samples and cause overfitting. Table~\ref{tab:test} shows the effect of inappropriate thresholds: increasing $t$ to $99.9\%$ on low-gap Llama3-v0.2 data sharply degrades performance, with repetitive generation observed, whereas reducing it to $95\%$ on high-gap Mistral data weakens performance.

		The CDF in Figure~\ref{fig:sub2} explains why $\xi$ should vary across datasets. Although Mistral and Llama3 both retain $99.9\%$ of samples below the threshold, their corresponding margins are 0.45 and 0.90. Thus, equal retained fractions can require substantially different $\xi$, reflecting the underlying reward-gap distributions rather than a universal numerical margin.

		\begin{table*}[t]
			\centering
			\caption{Quantiles of the reward gap $\Delta r$ distribution for different datasets.}
			\label{tab:qutile}
			\scriptsize
			\setlength{\tabcolsep}{5pt}
			\begin{tabular}{lccccccccccccc}
				\hline
				\multirow{2}{*}{Datasets} & \multicolumn{13}{c}{Quantiles of $\Delta r$  distribution ($\xi$)} \\ \cline{2-14}
				& 1\% & 5\% & 10\% & 25\% & 35\% & 45\% & 50\% & 75\% & 90\% & 95\% & 99\% & 100\% & $\xi$ in paper \\ \hline
				Mistral-7B        & -0.40 & -0.22 & -0.15 & -0.07 & -0.04 & -0.01 & 0.00 & 0.08 & 0.16 & 0.23 & 0.41 & 0.91 & 0.45 ($t=99.9\%$) \\
				Llama3            & -0.54 & -0.28 & -0.19 & -0.08 & -0.05 & -0.01 & 0.00 & 0.10 & 0.22 & 0.33 & 0.60 & 0.92 & 0.90 ($t=99.9\%$) \\
				Llama3-v2         & -0.51 & -0.26 & -0.18 & -0.09 & -0.05 & -0.01 & 0.00 & 0.11 & 0.25 & 0.35 & 0.64 & 0.92 & 0.35 ($t=95\%$) \\
				Gemma2             & -0.59 & -0.28 & -0.15 & -0.06 & -0.03 & -0.01 & 0.00 & 0.07 & 0.17 & 0.28 & 0.63 & 0.96 & 0.28 ($t=95\%$) \\ \hline
			\end{tabular}
		\end{table*}

		\begin{table}[!htbp]
			\centering
			\caption{Effect of overly strong or weak ratio margins. LC and WR are reported for each dataset and quantile threshold.}
			\label{tab:test}
			\small
			\begin{tabular}{@{}lrrr@{}}
				\toprule
				Dataset & $t$ & LC & WR \\
				\midrule
				Llama3-v0.2 (low-gap) & 0.95 & \textbf{57.5} & \textbf{50.5} \\
				& 0.999 & 19.2 & 20.1 \\
				Mistral (high-gap) & 0.95 & 29.3 & 30.3 \\
				& 0.999 & \textbf{33.2} & \textbf{36.5} \\
				\bottomrule
			\end{tabular}
		\end{table}

		\section{$\xi$-DPO on Verifiable Task}
		\label{sec:sql}
		\textbf{Data and evaluation.} We additionally evaluate Text-to-SQL~\cite{text2sql1,text2sql2,text2sql3} on Spider~\cite{spider}, using OmniSQL's~\cite{omnisql} preprocessed training and test splits. Prompts include database schemas; chosen responses contain the supplied chain-of-thought reasoning and gold SQL. We generate rejected responses with OmniSQL-32B to construct low-gap preference pairs. Execution accuracy compares the database execution results of generated and gold SQL. Each model is evaluated with a single inference pass, without voting or other accuracy-boosting strategies.
		
		\begin{table}[!htbp]
			\centering
			\caption{Execution accuracy on Spider test data.}
			\label{tab:sql}
			\small
			\setlength{\tabcolsep}{3pt}
			\begin{tabular}{@{}lrrrrr@{}}
				\toprule
				Method & Easy & Medium & Hard & Extra & All \\
				\midrule
				SFT & 64.9 & 44.0 & 52.5 & 47.6 & 51.0 \\
				SimPO & \textbf{88.1} & \uline{60.3} & \uline{69.3} & \textbf{62.7} & \uline{68.7} \\
				$\xi$-DPO & \uline{87.9} & \textbf{60.8} & \textbf{70.0} & \uline{62.5} & \textbf{69.0} \\
				\bottomrule
			\end{tabular}
		\end{table}
		
		\textbf{Training and results.} Starting from OmniSQL-7B, already supervised fine-tuned on large-scale Text-to-SQL data, we train SimPO and $\xi$-DPO on the constructed preference pairs. Other settings follow Appendix~\ref{sec:setup}; Equation~\eqref{eq:quntile} with $t=95\%$ gives $\xi=0.26$, while SimPO uses its original hyperparameter settings~\cite{simpo}. Table~\ref{tab:sql} reports all difficulty levels: $\xi$-DPO raises overall execution accuracy from 51.0\% to 69.0\%, compared with SimPO's 68.7\%, supporting its effectiveness beyond the original evaluation setting.

		\section{Proof of the Effect of $\beta$ on the Log-Sigmoid Objective Gradient}
		\label{sec:prof_first}
		Let $\Delta=|y_w|^{-1}\log\pi_\theta(y_w\mid x)-|y_l|^{-1}\log\pi_\theta(y_l\mid x)$, $z=\beta\Delta-\gamma$, and $\mathcal{J}=\log\sigma(z)$. Using $\sigma'(z)=\sigma(z)[1-\sigma(z)]$ and the chain rule,
		\begin{equation}
			\begin{aligned}
				\frac{\partial\mathcal{J}}{\partial z}
				&=\frac{\sigma'(z)}{\sigma(z)}=1-\sigma(z),\\
				\frac{\partial\mathcal{J}}{\partial\Delta}
				&=\beta[1-\sigma(\beta\Delta-\gamma)].
			\end{aligned}
		\end{equation}
		Thus, $\beta$ changes sample weighting as well as gradient scale: increasing it sharpens gradient decay for positive tail gaps while strengthening updates for negative-gap head samples. These coupled effects explain Figure~\ref{fig:distribution}. The corresponding negative-loss gradient is used in Appendix~\ref{sec:prof}.

		\section{$\xi$-DPO without LeakyReLU}
		\label{sec:worelu}
\begin{figure*}[t]
			\centering
			\begin{subfigure}[b]{0.48\textwidth}
				\centering
				\includegraphics[width=0.97\textwidth]{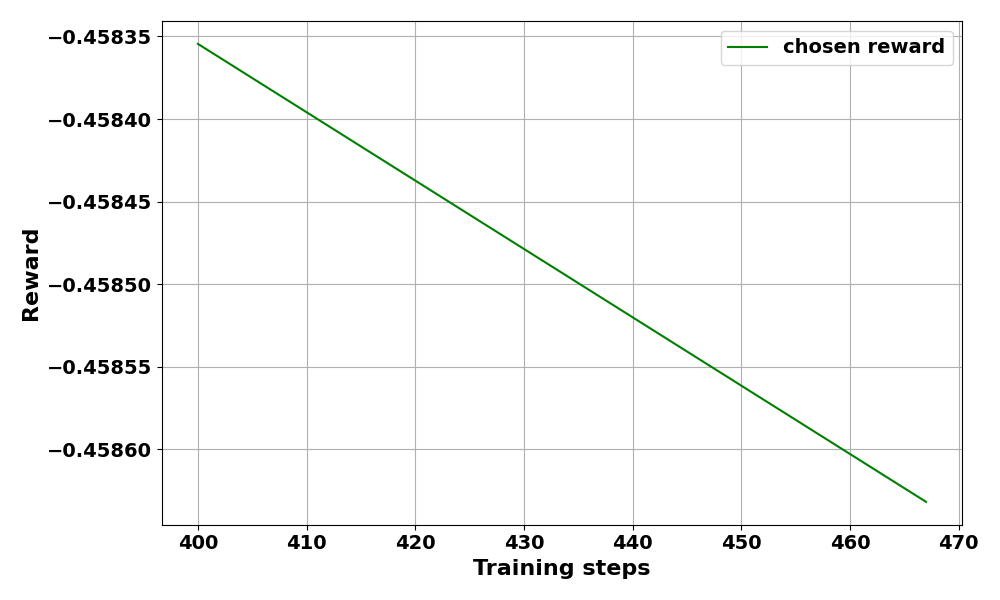}
				\caption{}
				\label{fig:chosen_bad}
			\end{subfigure}
			\hfill
			\begin{subfigure}[b]{0.48\textwidth}
				\centering
				\includegraphics[width=0.97\textwidth]{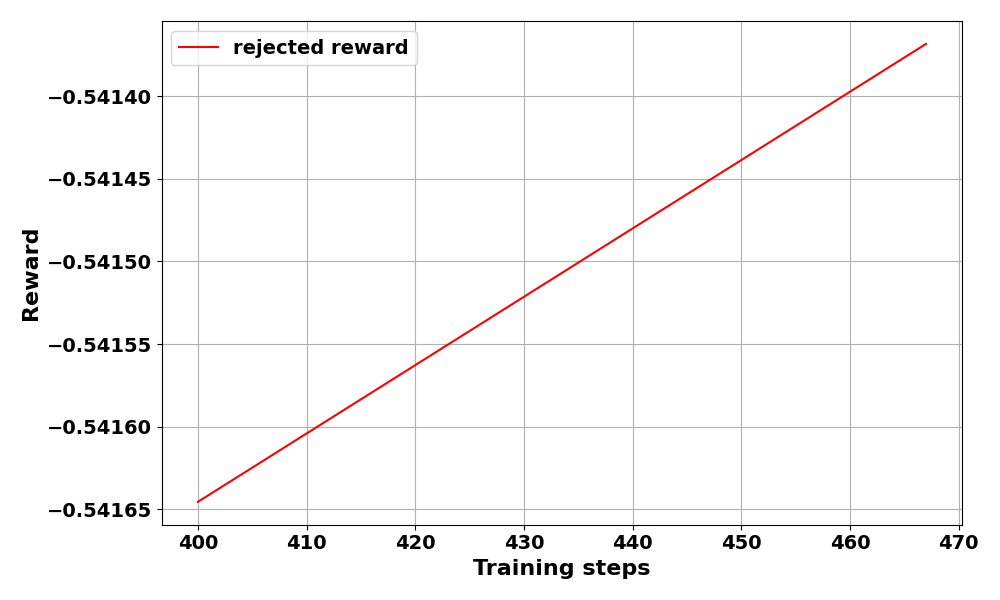}
				\caption{}
				\label{fig:rejected_bad}
			\end{subfigure}
			\caption{Reward curves of model training without LeakyReLU.}
			\label{fig:bad}
		\end{figure*}

		Figure~\ref{fig:bad} provides the reward curves discussed in Section~\ref{sec:na}.
		
		\section{Detailed experimental setup}
		\label{sec:setup}
		All experiments are conducted on four RTX PRO 6000 GPUs, using exactly the same hyperparameter settings as AlphaDPO\cite{aDPO}. The batch size is set to 128, and the learning rates are set to  $6.0 \times 10^{-7}$, $1.0 \times 10^{-6}$, and $8.0 \times 10^{-7}$ for Mistral-Instruct, Llama3-Instruct, and Gemma2, respectively. Specifically, on the Spider dataset, we adopt the same experimental configuration as Mistral, with a learning rate of $6.0 \times 10^{-7}$. The learning rate schedule is also kept consistent with AlphaDPO, using a cosine learning-rate schedule with a 10\% warm-up phase. The optimizer is Adam, also following AlphaDPO.

		\section{Preference optimization methods}
		The published objectives are given in the cited works; Table~\ref{tab:all_method} retains their hyperparameter settings and search ranges. It can be observed that although AlphaDPO\cite{aDPO} adopts an adaptive strategy for adjusting $\gamma$ it still requires an initial value of $\gamma$, while introducing an additional hyperparameter $\alpha$ and a reference model, which is involved in an additional computation. Meanwhile, for both SimPO and AlphaDPO, the search range of $\beta$ is relatively broad, making hyperparameter selection challenging in practical scenarios. In contrast, the proposed $\xi$-DPO has a simple formulation, and its hyperparameter can be determined through prior computation, making it more suitable for downstream applications.
		
		\begin{table}[t]
			
			\centering
			\caption{Reproduction results of the hyperparameter-free SimPER method on Mistral-Instruct.}
			\label{tab:simper_reproduction}
			\small
			\begin{tabular}{lcc}
				\toprule
				\multirow{2}{*}{Method} & \multicolumn{2}{c}{Mistral-Instruct (7B)} \\
				\cmidrule(lr){2-3}
				& LC(\%) & WR(\%) \\
				\midrule
				SimPER & 24.3 & 29.3 \\
				$\xi$-DPO & \textbf{33.2} & \textbf{36.5} \\
				\bottomrule
			\end{tabular}
			
		\end{table}
		
		The results in Table \ref{tab:main_results} are obtained using the official AlpacaEval 2 package\cite{alpacal} under the same inference configuration, where both the judge model and the baseline model are GPT-4-Turbo. To ensure a fair comparison with these results, we fine-tune Mistral using SimPER with the same learning rate and other hyperparameters, and evaluate the resulting model with GPT-4-Turbo. As shown in Table \ref{tab:simper_reproduction}, optimization without margin constraints performs poorly, highlighting the importance of the reward margin.

		\begin{table}[!htbp]
			\centering
			\caption{Hyperparameter settings and search ranges of preference optimization methods.}
			\label{tab:all_method}
			\small
			\setlength{\tabcolsep}{3pt}
			\begin{tabular}{@{}ll@{}}
				\toprule
				Method & Hyperparameters \\
				\midrule
				DPO~\cite{DPO} & $\beta\in\{0.01,0.05,0.1\}$ \\
				IPO~\cite{IPO} & $\tau\in\{0.01,0.1,0.5,1.0\}$ \\
				CPO~\cite{COT} & $\alpha=1.0,\ \beta\in\{0.01,0.05,0.1\}$ \\
				KTO~\cite{KTO} & \makecell[l]{$\lambda_l=\lambda_w=1.0$\\$\beta\in\{0.01,0.05,0.1\}$} \\
				ORPO~\cite{ORPO} & $\lambda\in\{0.1,0.5,1.0,2.0\}$ \\
				R-DPO~\cite{R-DPO} & \makecell[l]{$\alpha\in\{0.05,0.1,0.5,1.0\}$\\$\beta\in\{0.01,0.05,0.1\}$} \\
				SimPO~\cite{simpo} & \makecell[l]{$\beta\in\{2.0,4.0,6.0,8.0\}$\\$\gamma\in\{0.3,0.5,1.0,1.2,1.4,1.6\}$} \\
				AlphaDPO~\cite{aDPO} & \makecell[l]{$\beta\in[2.5,10.0],\ \gamma\in\{0.1,0.3,0.5\}$\\$\alpha\in\{10^{-2},5\!\times\!10^{-2},0.1,0.2\}$} \\
				SimPER~\cite{simper} & -- \\
				$\xi$-DPO & $t\in \{0.95,0.999\}$ \\
				\bottomrule
			\end{tabular}
		\end{table}
		
		\section{Theoretical Analysis of Objective Equivalence and Gradient}
		\label{sec:prof}
		
		\subsection{Analysis of Objective Equivalence}
		The logit transformation discussed with IPO~\cite{IPO} is strictly increasing, since $\operatorname{logit}'(p)=1/[p(1-p)]>0$ for $p\in(0,1)$. This preserves preference-probability ordering and is distinct from ratio normalization, whose purpose is to cancel $\beta$ and define a bounded margin.
		
		The order-preservation result for ratio normalization is proved in Section~\ref{sec:xi_definition}.

		\subsection{Analysis of Objective Gradient}
		We compare the gradients of SimPO and $\xi$-DPO. Gradient weighting is not unique to the proposed ratio objective. For example, the per-example SimPO loss is
		\begin{equation}
			\begin{aligned}
				L_i^{\mathrm{SimPO}}
				=
				-\log \sigma(\beta d_i-\gamma),
			\end{aligned}
		\end{equation}
		where $d_i=r_{w,i}-r_{l,i}$. Its gradient is
		\begin{equation}
			\begin{aligned}
				\nabla_\theta L_i^{\mathrm{SimPO}}
				=
				-\beta
				\left[
				1-\sigma(\beta d_i-\gamma)
				\right]
				\nabla_\theta d_i.
			\end{aligned}
		\end{equation}
		
		Thus, SimPO produces a sample-dependent weighted sum at the minibatch level, with the weight determined by the current reward gap. This is the mechanism analyzed in our paper as sigmoid-based sample filtering.
		
		For $\xi$-DPO, let
		\begin{equation}
			\begin{aligned}
				m_i
				&=
				\frac{d_i}{s_i},
				&
				s_i
				&=
				\operatorname{sg}
				\left(
				|r_{w,i}+r_{l,i}|
				\right)
				>0,
			\end{aligned}
		\end{equation}
		and let
		\begin{equation}
			\begin{aligned}
				L_i^{\xi\text{-DPO}}
				=
				\left[
				\phi(\xi-m_i)
				\right]^2,
			\end{aligned}
		\end{equation}
		where $\phi$ denotes LeakyReLU. The complete per-example gradient is
		\begin{equation}
			\begin{aligned}
				\nabla_\theta L_i^{\xi\text{-DPO}}
				=
				-
				\frac{
					2\phi(\xi-m_i)\phi'(\xi-m_i)
				}{
					s_i
				}
				\nabla_\theta d_i.
			\end{aligned}
		\end{equation}
		
		The minibatch gradient is a sample-dependent weighted sum. However, the factor $1/s_i$ is not introduced as an independently chosen importance weight. It follows directly from parameterizing the optimization target using the relative gap $m_i=d_i/s_i$ rather than the absolute gap $d_i$. For example, the reward pairs $(-1,-3)$ and $(-10,-30)$ have different absolute gaps, namely $2$ and $20$, but the same ratio gap:
		\begin{equation}
			\begin{aligned}
				\frac{-1-(-3)}{|-1+(-3)|}
				=
				\frac{-10-(-30)}{|-10+(-30)|}
				=
				0.5.
			\end{aligned}
		\end{equation}
		
		The ratio representation therefore treats pairs with the same relative chosen--rejected separation consistently, even when their absolute log-probability scales differ. Equivalently, it is invariant to a common positive scaling of both rewards within a pair.
		
		Accordingly, the induced weighting is the consequence of optimizing a relative-margin objective, rather than an arbitrary sample-reweighting heuristic. LeakyReLU further suppresses the contribution of examples whose ratio gap already exceeds the target margin.

	\end{document}